\documentclass[letterpaper, 10 pt, conference]{ieeeconf}  % Comment this line out if you need a4paper
\IEEEoverridecommandlockouts                              % This command is only needed if 
\usepackage{bm}
\usepackage{url}
\usepackage{graphicx} % for pdf, bitmapped graphics files
\usepackage{mathptmx} % assumes new font selection scheme installed
\usepackage[mathscr]{euscript}  
\usepackage{amsmath} % assumes amsmath package installed
\usepackage{amssymb}  % assumes amsmath package installed
\usepackage{balance}
\usepackage{multirow}
\usepackage[load=prefixed]{siunitx}
\usepackage{xcolor}
\usepackage{adjustbox}
\usepackage{array}
\usepackage{booktabs}

\newcommand{\secref}[1]{Sec.~\ref{#1}}

\renewcommand{\eqref}[1]{Eq.~(\ref{#1})}
\newcommand{\figref}[1]{Fig.~\ref{#1}}
\newcommand{\tabref}[1]{Tab.~\ref{#1}}

\DeclareMathAlphabet\mathbfcal{OMS}{cmsy}{b}{n}
\DeclareSIUnit\degm{deg/m}

\newcolumntype{P}[1]{>{\centering\arraybackslash}p{#1}}

\renewcommand{\baselinestretch}{0.97}
\newcolumntype{R}[2]{%
    >{\adjustbox{angle=#1,lap=\width-(#2)}\bgroup}%
    l%
    <{\egroup}%
}

\title{\LARGE \bf
Deep Auxiliary Learning for Visual Localization and Odometry
}

\author{Abhinav Valada$^*$ \and Noha Radwan$^*$ \and Wolfram Burgard% <-this % stops a space
\thanks{$^*$These authors contributed equally. All authors are with the Department of Computer Science, University of Freiburg, Germany. This work has partially been supported by the European
Commission under the grant number H2020-ICT-644227-FLOURISH.}% <-this % stops a space
}

\begin{document}

\onecolumn
{\Large

\noindent\textcopyright IEEE. Personal use of this material is permitted. Permission from IEEE must be obtained
for all other uses, in any current or future media, including reprinting/republishing this material
for advertising or promotional purposes, creating new collective works, for resale or redistribution
to servers or lists, or reuse of any copyrighted component of this work in other works.\\

%\linebreak
%
\noindent{Pre-print of article that will appear at the\\ \textbf{2018 IEEE International Conference on Robotics and Automation (ICRA 2018)}.}\\

\noindent{Please cite this paper as:}\\
A. Valada, N. Radwan, W. Burgard, "Deep Auxiliary Learning for Visual Localization and Odometry", \textit{IEEE International Conference on Robotics and Automation (ICRA)}, 2018.\\

\noindent{BibTex:}\\
\\
@inproceedings$\lbrace$valada18icra,\\
author = $\lbrace$Abhinav Valada and Noha Radwan and Wolfram Burgard$\rbrace$,\\
booktitle = $\lbrace$International Conference on Robotics and Automation (ICRA 2018)$\rbrace$,\\
organization = $\lbrace$IEEE$\rbrace$,\\
title = $\lbrace$Deep Auxiliary Learning for Visual Localization and Odometry$\rbrace$,\\
year = $\lbrace$2018$\rbrace$ \\
$\rbrace$
}
\twocolumn

\maketitle
\thispagestyle{empty}
\pagestyle{empty}

%%%%%%%%%%%%%%%%%%%%%%%%%%%%%%%%%%%%%%%%%%%%%%%%%%%%%%%%%%%%%%%%%%%%%%%%%%%%%%%%
\begin{abstract}
Localization is an indispensable component of a robot's autonomy stack that enables it to determine where it is in the environment, essentially making it a precursor for any action execution or planning. Although convolutional neural networks have shown promising results for visual localization, they are still grossly outperformed by state-of-the-art local feature-based techniques. In this work, we propose VLocNet, a new convolutional neural network architecture for 6-DoF global pose regression and odometry estimation from consecutive monocular images. Our multitask model incorporates hard parameter sharing, thus being compact and enabling real-time inference, in addition to being end-to-end trainable. We propose a novel loss function that utilizes auxiliary learning to leverage relative pose information during training, thereby constraining the search space to obtain consistent pose estimates. We evaluate our proposed VLocNet on indoor as well as outdoor datasets and show that even our single task model exceeds the performance of state-of-the-art deep architectures for global localization, while achieving competitive performance for visual odometry estimation. Furthermore, we present extensive experimental evaluations utilizing our proposed Geometric Consistency Loss that show the effectiveness of multitask learning and demonstrate that our model is the first deep learning technique to be on par with, and in some cases outperforms state-of-the-art SIFT-based approaches.
\end{abstract}

%%%%%%%%%%%%%%%%%%%%%%%%%%%%%%%%%%%%%%%%%%%%%%%%%%%%%%%%%%%%%%%%%%%%%%%%%%%%%%%%

\section{Introduction}
\label{sec:introduction}

Visual localization is a fundamental transdisciplinary problem and a crucial enabler for numerous robotics as well as computer vision applications, including autonomous navigation, Simultaneous Localization and Mapping (SLAM), Structure-from-Motion (SfM) and Augmented Reality (AR). More importantly, it plays a vital role when robots lose track of their location, or what is commonly known as the kidnapped robot problem. In order for robots to be safely deployed in the wild, their localization system should be robust to frequent changes in the environment; whether environmental changes such as illumination and seasonal appearance, dynamic changes such as moving vehicles and pedestrians, or structural changes such as constructions. 

Visual localization techniques can be broadly classified into two categories; topological and metric methods. Topological localization provides coarse estimates of the position, usually by dividing the map into a discretized set of locations and employing image retrieval techniques~\cite{arandjelovic2016netvlad, cummins2008fab, sunderhauf2015performance}. While this approach is well suited for large environments, the resulting location accuracy is bounded by the granularity of the discrete set. Metric localization approaches on the other hand, provide a 6-DoF metric estimate of the pose within the environment. Thus far, local feature-based approaches that utilize SfM information achieve state-of-the-art performance~\cite{sattler2017, valentin2015exploiting}. However, a critical drawback of these approaches is the decrease in speed and increase in complexity of finding feature correspondences as the size of the environment grows. Moreover, most approaches require a minimum
number of matches to be able to produce a pose estimate. This in turn causes pose estimation failures when there is large viewpoint changes, motion blur, occlusions or textureless environments.
%
% Moreover, as most approaches employ an n-point solver within a RANSAC loop, they require a minimum number of feature correspondences to be able to produce a reasonable pose estimate, which often results in failures when there is large viewpoint changes, motion blur, occlusions and textureless environments.

\begin{figure}
\centering
\includegraphics[width=\linewidth]{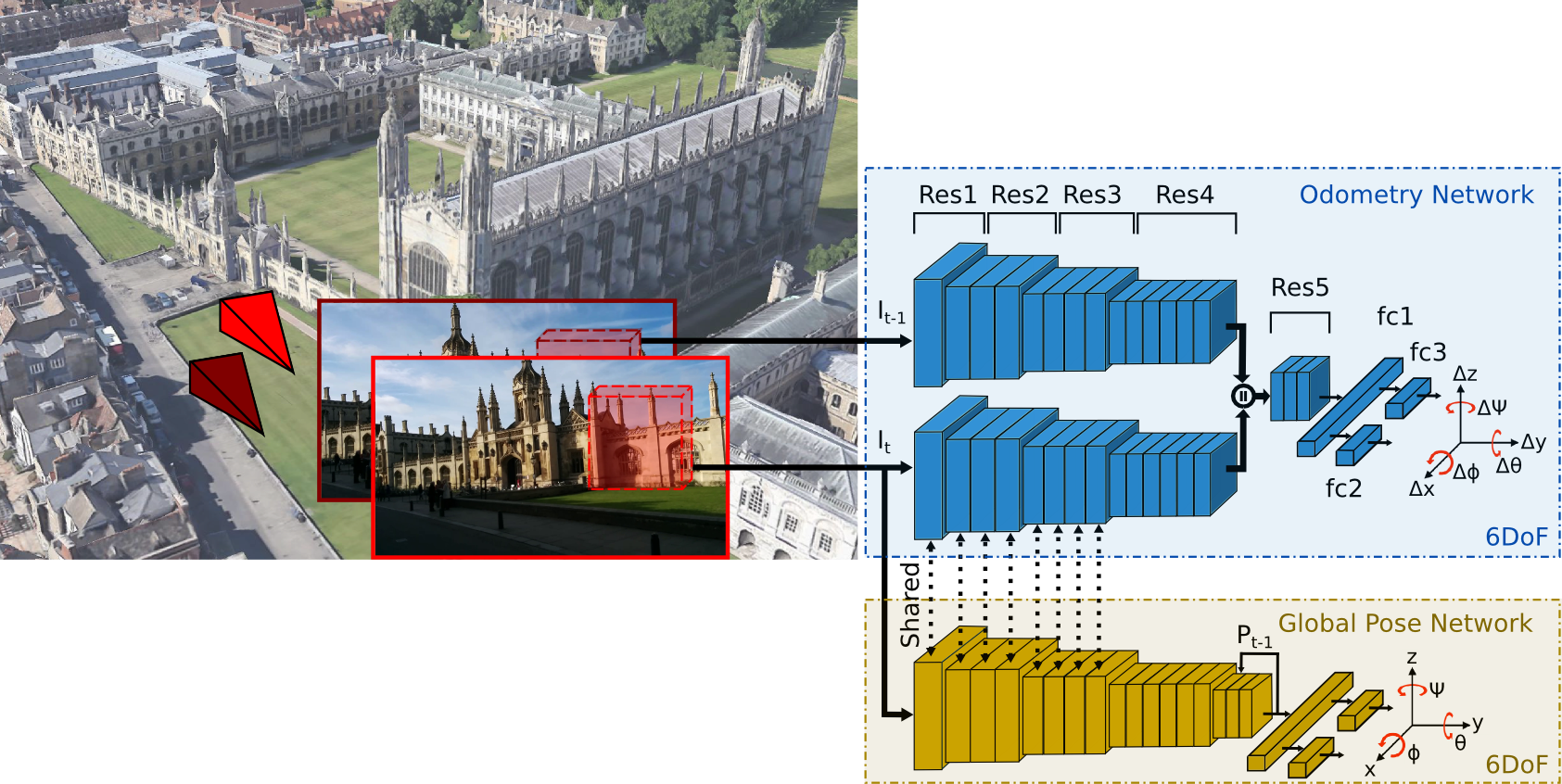}
\caption{\textbf{VLocNet: Multitask deep convolutional neural network for \mbox{6-DoF} visual localization and odometry.} Our network takes two consecutive monocular images as input and regresses the 6-DoF global pose and \mbox{6-DoF} odometry simultaneously. The global pose and odometry sub-networks incorporate hard parameter sharing and utilize our proposed Geometric Consistency Loss function that is robust to environmental aliasing. Online demo: \protect\url{http://deeploc.cs.uni-freiburg.de/}}
\label{fig:cover}
\vspace{-0.5cm}
\end{figure}

Inspired by the outstanding performance of Convolutional Neural Networks (CNNs) in a variety of tasks in various domains and with the goal of eliminating manual engineering of algorithms for feature selection, CNN architectures that directly regress the 6-DoF metric pose have recently been explored~\cite{kendall2015convolutional, walch16spatialstms, Clark2017VidLoc6V}. However, despite their ability to handle challenging perceptual conditions and effectively manage large environments, they are still unable to match the performance of state-of-the-art feature-based localization methods. This is partly due to their inability to internally model the 3D structural constraints of the environment while learning from a single monocular image. 

As CNN-based approaches become the de-facto standard for more robotics tasks, the need for multitask models becomes increasingly crucial. Moreover, from a robot's learning perspective, it is unlucrative and unscalable to have multiple specialized single-task models as they inhibit both inter-task and auxiliary learning. This has lead to a recent surge in research targeted towards frameworks for learning unified models for a range of tasks across different domains~\cite{yu3024, rahmatizadeh2017, bilenV17}. The goal of these multitask learning methods is to leverage similarities within task-specific features and exploit complementary features learned across different tasks, with the aim of mutual benefit. An evident advantage is the resulting compact model size in comparison to having multiple task-specific models. Auxiliary learning approaches on the other hand, aim at maximizing the prediction of a primary task by supervising the model to additionally learn a secondary task. For instance, in the context of localization, humans often describe their location to each other with respect to some reference landmark in the scene and giving their position relative to it. Here, the primary task is to localize and the auxiliary task is to be able to identify landmarks. Similarly, we can leverage the complementary relative motion information from odometry to constrict the search space while training the global localization model. However, this problem is non-trivial as we need to first determine how to structure the architecture to ensure the learning of this inter-task correlation and secondly, how to jointly optimize the unified model since different task-specific networks have different attributes and different convergence rates. 

%The idea of utilizing relative motion information to estimate the global pose closely resembles that of constructing an SfM model of the environment from sequential images and motion commands. 

In this work, we address the problem of global pose regression by simultaneously learning to estimate visual odometry as an auxiliary task. We propose the VLocNet architecture consisting of a global pose regression sub-network and a Siamese-type relative pose estimation sub-network. Our network based on the residual learning framework, takes two consecutive monocular images as input and jointly regresses the 6-DoF global pose as well as the 6-DoF relative pose between the images. We incorporate a hard parameter sharing scheme to learn inter-task correlations within the network and present a multitask alternating optimization strategy for learning shared features across the network. Furthermore, we devise a new loss function for global pose regression that incorporates the relative motion information during training and enforces the predicted poses to be geometrically consistent with respect to the true motion model. We present extensive experimental evaluations on both indoor and outdoor datasets comparing the proposed method to state-of-the-art approaches for global pose regression and visual odometry estimation. We empirically show that our proposed VLocNet architecture achieves state-of-the-art performance compared to existing CNN-based techniques. To the best of our knowledge, our presented approach is the first deep learning-based localization method to perform on par with local feature-based techniques. Moreover, our work is the first attempt to show that a joint multitask model can precisely and efficiently outperform its task-specific counterparts for global pose regression and visual odometry estimation.

%Results demonstrate the  efficacy of the learned model, achieving an improvement of $77.14\%$ and $52.03\%$ in translation and rotation respectively over current state-of-the-art deep learning methods.

%To summarize, we make the following contributions:
%\begin{itemize}
%\item We propose a new residual DCNN architecture for \mbox{6-DoF} global pose regression.
%\item We propose a new residual Siamese DCNN architecture for 6-DoF odometry estimation. 
%\item We propose a novel auxiliary learning framework with an alternating optimization strategy for jointly regressing the global pose and visual odometry. Our approach presents an efficient, and scalable alternative to learning task specific models.
%\item We devise a novel loss function that incorporates the relative motion information during training, enabling the network to predict pose estimates that consistent with the true motion model.
%\item We present comprehensive quantitative comparisons on the performance of our task specific networks as well as our joint multi-task network with DCNN-based approaches and state-of-the-art local feature-based techniques on two publicly available datasets.
%\item We provide concrete evidence on the efficacy of using deep auxiliary learning for global pose regression.
%\end{itemize}

\section{Related Work}
\label{sec:relatedWork}

There are numerous approaches that have been proposed for localization in the literature. In this section, we review some of the techniques developed thus far for addressing this problem, followed by a brief discussion on approaches for visual odometry estimation. 

\textbf{Sparse feature-based localization} approaches learn
a set of feature descriptors from training images and build a
codebook of 3D descriptors against which a query image is
matched~\cite{sattler2017, hao20123d}. To efficiently find
feature correspondences within the codebook, Shotton~\textit{et~al.}~\cite{shotton2013cvpr} and Valentin~\textit{et~al.}~\cite{valentin2015exploiting} train regression forests on 3D
scene data and use RANSAC to infer the final location of the query
image. Donoser~\textit{et al.} propose a
discriminative classification approach using random
ferns and demonstrate improved pose accuracy with faster run times~\cite{donoser2014discriminative}. Despite the
accurate pose estimates provided by these methods, the overall run-time
depends on the size of the 3D model and the number of feature correspondences found.
The approach presented in this paper does not suffer from these
scalability issues as the learned model is independent of the size 
of the environment. Moreover, since it does not involve any expensive matching
algorithm, it has a time complexity of $\mathbfcal{O}(1)$.

\textbf{Deep learning-based localization:} PoseNet~\cite{kendall2015convolutional} was the first approach to utilize DCNNs to address the metric localization problem. The authors further extended this work by using a Bayesian CNN implementation to estimate the uncertainty of
the predicted pose~\cite{kendall2015modelling}. Concurrently,
Walch~\textit{et~al.}~\cite{walch16spatialstms} and Clark~\textit{et~al.}~\cite{Clark2017VidLoc6V} propose DCNNs with Long-Short Term
Memory (LSTM) units to avoid overfitting
while still selecting the most useful feature correlations. Contrary
to these approaches and inspired by semantic segmentation
architectures, Melekhov~\textit{et al.} introduce the HourglassPose~\cite{MelekhovYKR17}
network that utilizes a symmetric encoder-decoder
architecture followed by a regressor to estimate the camera pose. 
%To alleviate the need for a large amount of training data,
%Naseer~\textit{et al.}~\cite{naseer17iros} and Wu~\textit{et
%  al.}~\cite{wu2017} propose augmentation strategies to increase the
%3D pose coverage of images. Additionally, Wu~\textit{et al.} propose
%to use the Euler6 representation to counteract ambiguities of
%quaternions in representing angles. 
In order to provide a more robust approach to balance both the translational and rotational components
in the loss term, the commonly employed fixed weight regularizer was replaced with learnable parameters in~\cite{kendall2017geometric}. The authors also introduced a loss function based on the geometric reprojection error that does not require balancing of the pose components, but it often has difficulty in converging. More recently, Laskar~\textit{et al.} proposed a learning procedure that decouples feature learning and pose estimation, closely resembling feature-based localization approaches~\cite{laskar2017camera}. Unlike most of the aforementioned approaches that utilize the Euclidean loss for pose regression, we propose a novel loss function that incorporates motion information while training to learn poses that are consistent with the previous prediction.

\textbf{Visual Odometry:} Another closely related problem in robotics is estimating the incremental motion of the robot using only sequential camera images. In one of the earlier approaches, Konda~\textit{et al.}~\cite{visapp15} adopt a classification approach to the problem, where a CNN with a $\mathsf{softmax}$ layer is used to infer the relative transformation between two images using a prior set of discretized velocities and directions. Another approach is proposed by Nicholai~\textit{et~al.}~\cite{NicolaiRSSw2016}, in which they combine both image and LiDAR information to estimate the relative motion between two frames. They project the point cloud on the 2D
image and feed this information to a neural network which estimates the visual odometry. Mohanty~\textit{et al.}~\cite{MohantyDeepVO} propose a Siamese AlexNet-based architecture called
DeepVO, in which the translational and rotational components are regressed through an L2-loss layer with equal weight values. In similar work, Melekhov~\textit{et~al.}~\cite{melekhov2017cnnBspp} add a weighting term to balance both the translational and rotational components of the loss, which yields
an improvement in the predicted pose. Additionally, they use a spatial pyramid pooling layer in their architecture which renders their approach robust to varying input image resolutions. Inspired by the
success of residual networks in various visual recognition tasks, we propose a Siamese-type two stream architecture built upon the ResNet-50~\cite{He2015} model for visual odometry estimation.

Contrary to the task-specific approaches presented above where individual models are
trained for global pose regression and visual odometry estimation, we
propose a joint end-to-end trainable architecture that simultaneously
regresses the 6-DoF global pose and relative motion as an auxiliary output.
By jointly learning both tasks, our approach is robust to environmental 
aliasing by utilizing previous pose and relative motion information, 
thereby combining the advantages of both local feature and deep learning-based
localization methods. Moreover, by sharing features across different scales, 
our proposed model significantly outperforms the state-of-the-art in CNN-based 
localization while achieving competitive performance for visual odometry estimation.

\section{Deep Pose Regression}
\label{sec:deepPoseReg}

%In this section, we describe in detail our proposed \mbox{VLocNet} architecture for regressing global poses and simultaneously learning to regress relative motion between two camera frames using only pairs of RGB images. 
The primary goal of our architecture is to precisely estimate the global pose by minimizing our proposed Geometric Consistency Loss function, which in turn constricts the search space using the relative motion between two consecutive frames. We formulate this problem in the context of auxiliary learning with the secondary goal of estimating the relative motion. The features learned for relative motion estimation are then leveraged by the global pose regression network to learn a more distinct representation of the scene. More specifically, our architecture consists of a three-stream neural network; a global pose regression stream and a Siamese-type double-stream for odometry estimation. An overview of our proposed VLocNet architecture is shown in Fig.~\ref{fig:cover}. Given a pair of consecutive monocular images $(I_t, I_{t-1})$, our network predicts both the global pose $\mathbf{p}_{t}=[\mathbf{x}_{t}, \mathbf{q}_{t}]$ and the relative pose $\mathbf{p}_{t, t-1}=[\mathbf{x}_{t, t-1}, \mathbf{q}_{t, t-1}]$ between the input frames, where \mbox{$\mathbf{x} \in \mathbb{R}^3$} denotes the translation and $\mathbf{q} \in \mathbb{R}^4$ denotes the rotation in quaternion representation. For ease of notation, we assume that the quaternion outputs of the network have been normalized a priori. The input to the Siamese streams are the images $I_t$, $I_{t-1}$, while the input to the global pose stream is $I_t$. 
%Unlike conventional Siamese architectures, we do not share features across the two temporal streams, instead we share features across the global pose stream and the Siamese stream that takes $I_t$ as input, upto a certain depth. As our images do not encompass any spatially centered structure, sharing features across these streams is a viable solution to facilitate auxiliary learning and joint optimization of the task specific sub-networks. Furthermore, we employ a novel loss function that enforces geometric consistency within the predicted global poses by utilizing the relative pose estimates while training. 
In the remainder of this section, we present the constituting parts of our VLocNet architecture along with how the joint optimization is carried out.

\subsection{Global Pose Regression}

In this section, we describe the architecture of our global pose sub-network, which given an input image $I_t$ and a previous predicted pose $\hat{\mathbf{p}}_{t-1}$, predicts the 7-dimensional pose $\hat{\mathbf{p}}_t$. Similar to previous works~\cite{kendall2015convolutional,walch16spatialstms}, $\mathbf{p}$ is defined relative to an arbitrary global reference frame. %and we chose the quaternion representation for orientation due to the ease of mapping them back to rotations. This however causes ambiguity as $\mathbf{q}$ and $\mathbf{{-q}}$ represent the same rotation.
%comes at the cost of having to account for ambiguity where $\mathbf{q}$ and $\mathbf{{-q}}$ represent the same rotation.

\subsubsection{Network Architecture}

To estimate the global pose, we build upon the ResNet-50~\cite{He2015} architecture with the following modifications. The structure of our network is similar to ResNet-50 truncated before the last average pooling layer. The architecture is comprised of five residual blocks with multiple residual units, where each unit has a bottleneck architecture consisting of three convolutional layers in the following order: $1\times1$ convolution, $3\times3$ convolution, $1\times1$ convolution. Each of the convolutions is followed by batch normalization, scale and Rectified Linear Unit (ReLU). We modify the standard residual block structure by replacing ReLUs with Exponential Linear Units (ELUs)~\cite{clevert2015fast}. ELUs help in reducing the bias shift in the neurons, in addition to avoiding the vanishing gradient and yield faster convergence. We replace the last average pooling layer with global average pooling and subsequently add three inner-product layers, namely \textit{fc1}, \textit{fc2} and \textit{fc3}. The first inner-product layer \textit{fc1} is of dimension $1024$ and the following two inner-product layers are of dimensions $3$ and $4$, for regressing the translation $\mathbf{x}$ and rotation $\mathbf{q}$ respectively. Our proposed Geometric Consistency Loss, detailed in Sec.~\ref{sec:gcloss}, ensures that the predicted pose is consistent with that obtained by accumulating the relative motion to the previous pose. Therefore, we feed the previous pose (groundtruth pose during training and predicted pose during evaluation) to the network so that it can better learn about spatial relations of the environment. We do not incorporate recurrent units into our network as our aim in this work is to localize only using consecutive monocular images and not rely on long-term temporal features. We first feed the previous pose to an inner-product layer \textit{fc4} of dimension $D$ and reshape its output to $H \times W \times C$, which corresponds in shape to the output of the last residual unit before the downsampling stage. Both tensors are then concatenated and fed to the subsequent residual unit. In total, there are four downsampling stages in our network and we experiment with fusing at each of these stages in Sec.~\ref{sec:vlocEvaluation}.

\subsubsection{Geometric Consistency Loss}
\label{sec:gcloss}

Learning both translational and rotational pose components with the same loss function is inherently challenging due to the difference in scale and units between both the quantities. \eqref{eq:standardLossx} and \eqref{eq:standardLossq} describe the loss function for regressing the translational and rotational components in the Euclidean space.
\begin{align}\label{eq:standardLossx}
\mathcal{L}_{x}(I_{t}) &:=  \left \| \mathbf{x}_{t} - \hat{\mathbf{x}}_{t}  
\right \|_\gamma \\ \label{eq:standardLossq}
\mathcal{L}_{q}(I_{t}) &:= \left 
\| \mathbf{q}_{t} - \hat{\mathbf{q}}_{t}\right \|_\gamma ,
\end{align}
where $\mathbf{x}_t$ and $\mathbf{q}_t$ denote the ground-truth translation and rotation components, $\hat{\mathbf{x}}_{t}$ and $\hat{\mathbf{q}}_{t}$ denote their predicted counterparts and $\gamma$ refers to the $L^\gamma$-norm. In this work, we use the $L^2$ Euclidean norm. Previous work has shown that the performance of a model trained to jointly regress the position and orientation, outperforms two separate models trained for each task~\cite{kendall2017geometric}. Therefore, as the loss function is required to learn both the position and orientation, a weight regularizer $\beta$ is used to balance each of the loss terms. We can represent this loss function as:
\begin{equation}\label{eq:betaLoss}
  \mathcal{L}_{\beta}(I_{t}) := \mathcal{L}_{x}(I_{t}) + \beta \mathcal{L}_{q}(I_{t}).
\end{equation}

Although initial work~\cite{kendall2015convolutional,walch16spatialstms,wu2017,naseer17iros} has shown that by minimizing this function, the network is able to learn a valid pose regression model, it suffers from the drawback of having to manually tune the hyperparameter $\beta$ for each new scene in order to achieve reasonable results. To counteract this problem, recently~\cite{kendall2017geometric} learnable parameters were introduced to replace $\beta$. The resulting loss function is:  
\begin{equation}\label{eq:sigmaLoss}
 \mathcal{L}_{s}(I_{t}) := \mathcal{L}_{x}(I_{t}) \exp({-\hat{s}}_{x}) + \hat{s}_{x} + \mathcal{L}_{q}(I_{t}) \exp({-\hat{s}}_{q}) + \hat{s}_{q},
\end{equation}
where $\hat{s}_x$ and $\hat{s}_q$ are the two learnable variables. Each variable acts as a weighting for the respective component in the loss function. Although this formulation overcomes the problem of having to manually select a $\beta$ value for each scene, it does not ensure that the estimated poses are consistent with the previous motion.

As a solution to this problem, we propose a novel loss function that incorporates previous motion information, thereby producing consistent pose estimates. We introduce an additional constraint which bootstraps the loss function by penalizing pose predictions that contradict the relative motion. More precisely, in addition to the loss term shown in~\eqref{eq:sigmaLoss}, we enforce that the difference between $\mathbf{\hat{p}}_t$ and $\mathbf{\hat{p}}_{t-1}$ be as close to the groundtruth relative motion $\mathbf{p}_{t, t-1}$ as possible. We use $\mathcal{R}_{x}(I_{t})$ and $\mathcal{R}_{q}(I_{t})$ to denote the relative motion between the current image $I_{t}$ and the previous predicted pose $\hat{\mathbf{p}}_{t-1}$ as:
\begin{align}\label{eq:relativex}
\mathcal{R}_{x}(I_{t}) &:= \hat{\mathbf{x}}_{t} - \hat{\mathbf{x}}_{t-1}  \\ 
\mathcal{R}_{q}(I_{t}) &:=  \hat{\mathbf{q}}_{t-1}^{{-1}} \hat{\mathbf{q}}_{t}.  \label{eq:relativeq}
%\frac{\mathbf{q}_{t}}{\left \| \mathbf{q}_{t} \right \|}- \mathbf{q}_{t-1} \label{eq:relativeq}
\end{align}

The components from~\eqref{eq:relativex} and~\eqref{eq:relativeq} compute the relative motion in terms of the network's predictions. We integrate these components into an odometry loss term to minimize the variance between the predicted poses. The corresponding odometry loss can be formulated as:
\begin{align}
\mathcal{L}_{x_{odom}} (I_{t}) &:= \left \| \mathbf{x}_{t, t-1} - \mathcal{R}_{x}(I_{t}) \right \|_\gamma \\
\mathcal{L}_{q_{odom}} (I_{t}) &:= \left \| \mathbf{q}_{t, t-1} - \mathcal{R}_{q}(I_{t}) \right \|_\gamma,
\end{align}
where $\mathcal{L}_{x_{odom}}$ computes the difference between the ground-truth relative translational motion and its predicted counterpart, while $\mathcal{L}_{q_{odom}}$ computes a similar difference for the rotational component. We combine both the odometry loss terms with the loss function from~\eqref{eq:sigmaLoss}, thereby minimizing:
\begin{multline}\label{eq:custLoss}
\mathcal{L}_{Geo} (I_{t}) := \left( \mathcal{L}_{x}(I_{t}) + \mathcal{L}_{x_{odom}} (I_{t}) \right) \exp({-\hat{s}}_{x}) \\ + \hat{s}_{x} + 
 \left( \mathcal{L}_{q}(I_{t}) + \mathcal{L}_{q_{odom}} (I_{t}) \right) \exp({-\hat{s}}_{q}) + \hat{s}_{q}.
\end{multline}

We hypothesize that, by utilizing this relative motion in the loss function, the resulting trained model is more robust to perceptual aliasing within the environment.

\subsection{Visual Odometry}

In order to integrate motion specific features in our global pose regression network, we train an auxiliary network to regress the 6-DoF relative pose from the images $(I_t, I_{t-1})$. We do so by constructing a two stream Siamese-type network also based on the ResNet-50 architecture. We concatenate features from the two individual streams of ResNet-50 truncated before the last downsampling stage (\textit{Res5}). We then pass these concatenated feature maps to the last three residual units, followed by three inner-product layers, similar to our global pose regression network. We minimize the following loss function:
\begin{multline}\label{eq:voLoss}
 \mathcal{L}_{vo}(I_{t}, I_{t-1}) := \mathcal{L}_{x}(I_{t}, I_{t-1}) \exp({-\hat{s}}_{x})\\  + \hat{s}_{x} + \mathcal{L}_{q}(I_{t}, I_{t-1}) \exp({-\hat{s}}_{q}) + \hat{s}_{q}.
\end{multline}

With a slight abuse of notation, we use $\mathcal{L}_{x}(I_{t}, I_{t-1})$ to refer to the $L^2$ Euclidean loss in the translational component of the visual odometry and $\mathcal{L}_{q}(I_{t}, I_{t-1})$ for the rotational component. Similar to our approach used for the global pose regression, we additionally learn two weighting parameters to balance the loss between both components. We detail the training procedure in \secref{sec:training}.

\subsection{Deep Auxiliary Learning}

The idea of jointly learning both the global pose and visual odometry stems from the inherent similarities across both tasks in the feature space. More importantly, sharing features across both networks can enable a competitive and collaborative action as each network updates its own weights during backpropagation in an attempt to minimize the distance to the groundtruth pose. This symbiotic action introduces additional regularization while training, thereby avoiding overfitting. Contrary to the approaches that use a two stream shared Siamese network for visual odometry estimation, we do not share weights between the two temporal streams, rather we share weights between the stream that takes the image $I_t$ from the current timestep as input and the global pose regression stream. By learning separate discriminative features in each timestep before learning the correlation between them, the visual odometry network is able to effectively generalize to challenging corner cases containing motion blur and perceptual aliasing. The global pose regression network also benefits from this feature sharing, as the shared weights are pulled more towards areas of the image from which the relative motion can be easily estimated.
%as the visual odometry network pulls the shared weights more towards areas of the image from which the relative motion can be easily estimated. 
%This has a tremendous impact on the accuracy of the predicted global pose in multiple scenarios. Consider the following situation in which the network attempts to estimate the pose of an image in a textureless structurally symmetric environment. Due to the presence of environmental aliasing, the accuracy of the predicted pose can be substantially lower when compared to predictions in an environment with abundant structural variations. However, by jointly training both networks using hybrid hard parameter sharing, the global pose regression network can leverage the relative motion features from the odometry stream, thereby producing more accurate location estimates.

While sharing features across multiple networks can be inferred as a form of regularization, it is not clear a priori for how many layers should we maintain a shared stream. Sharing only a few initial layers does not have any additive benefit to either network, as early layers learn very generic feature representations. On the other hand, maintaining a shared stream too deep into the network can negatively impact the performance of both tasks, since the features learned at the stages towards the end are more task specific. In this work, we studied the impact of sharing features across both sub-networks and experimented with varying the amount of feature sharing. We elaborate on these experiments in~\secref{sec:auxEvaluation}. Another critical aspect of auxiliary learning is how the optimization is carried out. We detail our optimization procedure in~\secref{sec:training}. Finally, during inference, the joint model can be deployed as a whole or each sub-network individually, since the relative pose estimates are only used in the loss function and there is no inter-network dependency in terms of concatenating or adding features from either sub-networks. This gives additional flexibility at the time of deployment compared to architectures that have cross-connections or cross-network fusion.

\section{Experimental Evaluation}
\label{sec:reults}

In this section, we present results using our proposed VLocNet architecture in comparison to the state-of-the-art on both indoor and outdoor datasets, followed by detailed analysis on the architectural decisions and finally, we demonstrate the efficacy of learning visual localization models along with visual odometry as an auxiliary task.

\subsection{Evaluation Datasets}
\label{sec:datasets}

\begin{table*}
\footnotesize 
\centering
\caption{Comparison of median localization error of VLocNet with existing CNN models on the 7-Scenes dataset.}
\label{tab:7scenesCompGP}
\begin{tabular}{p{1.2cm}p{1.3cm}p{1.3cm}p{1.3cm}p{1.2cm}p{1.35cm}p{1.35cm}p{1.3cm}p{1.55cm} | p{1.6cm}}
\hline\noalign{\smallskip}
Scene & PoseNet~\cite{kendall2015convolutional} & Bayesian PoseNet~\cite{kendall2015modelling} & LSTM-Pose~\cite{walch16spatialstms} & VidLoc~\cite{Clark2017VidLoc6V} & Hourglass-Pose~\cite{MelekhovYKR17} & BranchNet \cite{wu2017} & PoseNet2 \cite{kendall2017geometric} & NNnet~\cite{laskar2017camera} & VLocNet (Ours) \\
\noalign{\smallskip}\hline\hline\noalign{\smallskip}
Chess & $0.32\meter,\, 8.12\degree$ & $0.37\meter,\, 7.24\degree$ & $0.24\meter,\, 5.77\degree$ & $0.18\meter,\, \text{N/A}$ & $0.15\meter,\, 6.53\degree$ & $0.18\meter,\, 5.17\degree$ & $0.13\meter,\, 4.48\degree$  & $0.13\meter,\, 6.46\degree$ & $\mathbf{0.036\meter,\, 1.71\degree}$ \\
Fire & $0.47\meter,\, 14.4\degree$ & $0.43\meter,\, 13.7\degree$ & $0.34\meter,\, 11.9\degree$ & $0.26\meter,\, \text{N/A}$ & $0.27\meter,\, 10.84\degree$ & $0.34\meter,\, 8.99\degree$ & $0.27\meter,\, 11.3\degree$  & $0.26\meter,\, 12.72\degree$  & $\mathbf{0.039\meter,\, 5.34\degree}$ \\
Heads & $0.29\meter,\, 12.0\degree$ & $0.31\meter,\, 12.0\degree$ & $0.21\meter,\, 13.7\degree$ & $0.14\meter,\, \text{N/A}$ & $0.19\meter,\, 11.63\degree$ & $0.20\meter,\, 14.15\degree$ & $0.17\meter,\, 13.0\degree$  & $0.14\meter,\, 12.34\degree$ & $\mathbf{0.046\meter, 6.64\degree}$ \\
Office & $0.48\meter,\, 7.68\degree$ & $0.48\meter,\, 8.04\degree$ & $0.30\meter,\, 8.08\degree$ & $0.26\meter,\, \text{N/A}$ & $0.21\meter,\, 8.48\degree$ & $0.30\meter,\, 7.05\degree$ & $0.19\meter,\, 5.55\degree$  & $0.21\meter,\, 7.35\degree$  & $\mathbf{0.039\meter,\, 1.95\degree}$ \\
Pumpkin & $0.47\meter,\, 8.42\degree$ & $0.61\meter,\, 7.08\degree$ & $0.33\meter,\, 7.00\degree$ &  $0.36\meter,\, \text{N/A}$ & $0.25\meter,\, 7.01\degree$ & $0.27\meter,\, 5.10\degree$ & $0.26\meter,\, 4.75\degree$  & $0.24\meter,\, 6.35\degree$ & $\mathbf{0.037\meter,\, 2.28\degree}$ \\
RedKitchen & $0.59\meter,\, 8.64\degree$ & $0.58\meter,\, 7.54\degree$ & $0.37\meter,\, 8.83\degree$ & $0.31\meter,\, \text{N/A}$ & $0.27\meter,\, 10.15\degree$ & $0.33\meter,\, 7.40\degree$ &  $0.23\meter,\, 5.35\degree$  & $0.24\meter,\, 8.03\degree$ & $\mathbf{0.039\meter,\, 2.20\degree}$ \\
Stairs & $0.47\meter,\, 13.8\degree$ & $0.48\meter,\, 13.1\degree$ & $0.40\meter,\, 13.7\degree$ & $0.26\meter,\, \text{N/A}$ & $0.29\meter,\, 12.46\degree$ & $0.38\meter,\, 10.26\degree$ & $0.35\meter,\, 12.4\degree$  & $0.27\meter,\, 11.82\degree$ & $\mathbf{0.097\meter,\, 6.48\degree}$ \\
\noalign{\smallskip}\hline\noalign{\smallskip}
Average & $0.44\meter,\, 10.4\degree$ & $0.47\meter,\, 9.81\degree$ & $0.31\meter,\, 9.85\degree$ & $0.25\meter,\, \text{N/A}$ & $0.23\meter,\, 9.53\degree$ & $0.29\meter,\, 8.30\degree$ & $0.23\meter,\, 8.12\degree$ & $0.21\meter,\, 9.30\degree$ & $\mathbf{0.048\meter,\, 3.80\degree}$ \\
\noalign{\smallskip}\hline\noalign{\smallskip}
\end{tabular}
\vspace{-0.4cm}
\end{table*}

We evaluate VLocNet on two publicly available datasets; Microsoft 7-Scenes~\cite{shotton2013cvpr} and Cambridge Landmarks~\cite{kendall2015convolutional}. We use the original train and test splits provided by all the datasets to facilitate comparison and benchmarking.

\subsubsection*{\textbf{Microsoft 7-Scenes}} is a dataset comprised of RGB-D images collected from seven different scenes in an indoor office environment~\cite{shotton2013cvpr}. The images were collected with a handheld Kinect RGB-D camera and the groundtruth poses were extracted using KinectFusion~\cite{shotton2013cvpr}. The images were captured at resolution of $640 \times 480$ pixels and each scene contains multiple sequences recorded in a room. Each sequence was recorded with different camera motions in the presence of motion blur, perceptual aliasing and textureless features in the room, thereby making it a popular dataset for relocalization and tracking. 

\subsubsection*{\textbf{Cambridge Landmarks}} provides images collected from five different outdoor scenes around the Cambridge University~\cite{kendall2015convolutional}. The images were captured using a smartphone at a resolution of $1920 \times 1080$ pixels while walking in different trajectories and pose labels were computed using an SfM method. %The images provided to the SfM pipeline were subsampled in time at $2\hertz$ with a distance of approximately $1\meter$ between each pose. Due to the nature with which the data was collected, the images exhibit high degree of noise, different weather and lighting conditions. 
The dataset exhibits substantial clutter caused by pedestrians, cyclists and moving vehicles, making it challenging for urban relocalization.

\subsection{Network Training}
\label{sec:training}

In order to train our network on different datasets, we rescale the images maintaining the aspect ratio such that the shorter side is of length 256 pixels. We calculate the pixel-wise mean for each of the scenes in the datasets and subtract them with the input images. We experimented with augmenting the images using pose synthesis~\cite{wu2017} and synthetic view synthesis~\cite{naseer17iros}, however they did not yield any performance gains, rather in some cases they negatively affected the pose accuracy. We found that using random crops of $224\times224$ pixels acts as a better regularizer helping the network generalize better in comparison to synthetic augmentation techniques while saving preprocessing time. For evaluations, we use the center crop of the images.

We use the Adam solver for optimization with $\beta_1=0.9, \beta_2=0.999$ and $\epsilon=10^{-10}$. We train the network with an initial learning rate of $\lambda_0 = 10^{-4}$ with a mini-batch size of $32$ and a dropout probability of 0.2. Details regarding the specific $\hat{s}_{x}$ and $\hat{s}_{q}$ values used for our Geometric Consistency Loss function are covered in \secref{sec:vlocEvaluation}. In order to learn a unified model and to facilitate auxiliary learning, we employ different optimization strategies that allow for efficient learning of shared features as well as task-specific features, namely alternate training and joint training. In alternate training we use a separate optimizer for each task and alternatively execute each task optimizer on the task-specific loss function, thereby allowing synchronized transfer of information from one task to the other. This instills a form of hierarchy into the tasks, as the odometry sub-network improves the estimate of its relative poses, the global pose network in turn uses this estimate to improve its prediction. It is often theorized that this enforces commonality between the tasks. The disadvantage of this approach is that a bias in the parameters is introduced by the task that is optimized second. In joint training on the other hand, we add each of the task-specific loss functions and use a single optimizer to train the sub-networks at the same time. The advantage of this approach is that the tasks are trained in a way that they maintain the individuality of their functions, but as each of our tasks is of different units and scale, the task with the larger scale often dominates the training.

We experiment with bootstrapping the training of VLocNet with different weight initializations for each of the aforementioned optimization schemes. Results from this experiment are discussed in~\secref{sec:auxEvaluation}. Using the principle of transfer learning, we trained the individual models by initializing all the layers up to the global pooling layer with the weights of ResNet-50 pretrained on ImageNet~\cite{ILSVRC15} and we used Gaussian initialization for the remaining layers. We use the TensorFlow~\cite{tensorflow2015} deep learning library and all the models were trained on a NVIDIA Titan X GPU for a maximum of 120,000 iterations, which approximately took 15 hours.

\subsection{Comparison with the State-of-the-art}
\label{sec:comparisonSota}

\begin{table*}
\footnotesize 
\centering
\caption{Comparison of median localization error of VLocNet with existing CNN models on the Cambridge Landmarks dataset.}
\label{tab:CLComp}
\begin{tabular}{p{2cm}p{1.5cm}p{1.5cm}p{1.5cm}p{1.5cm}p{1.7cm}|p{1.7cm}}
\hline\noalign{\smallskip}
Scene & PoseNet~\cite{kendall2015convolutional} & Bayesian PoseNet~\cite{kendall2015modelling} & SVS-Pose~\cite{naseer17iros} & LSTM-Pose~\cite{walch16spatialstms} & PoseNet2~\cite{kendall2017geometric} & VLocNet (Ours) \\
\noalign{\smallskip}\hline\hline\noalign{\smallskip}
King's College & $1.92\meter,\, 5.40\degree$ & $1.74\meter,\, 4.06\degree$ & $1.06\meter,\, 2.81\degree$ & $0.99\meter,\, 3.65\degree$ & $0.88\meter,\, 1.04\degree$ & $\mathbf{0.836\meter},\, 1.419\degree$ \\
Old Hospital & $2.31\meter,\, 5.38\degree$ & $2.57\meter,\, 5.14\degree$ & $1.50\meter,\, 4.03\degree$ & $1.51\meter,\, 4.29\degree$ & $3.20\meter,\, 3.29\degree$ & $\mathbf{1.075\meter,\, 2.411\degree}$ \\
Shop Facade & $1.46\meter,\, 8.08\degree$ & $1.25\meter,\, 7.54\degree$  & $0.63\meter,\, 5.73\degree$ & $1.18\meter,\, 7.44\degree$ & $0.88\meter,\, 3.78\degree$ & $\mathbf{0.593\meter,\, 3.529\degree}$ \\
St Mary's Church & $2.65\meter,\, 8.46\degree$ & $2.11\meter,\, 8.38\degree$ & $2.11\meter,\, 8.11\degree$ & $1.52\meter,\, 6.68\degree$  & $1.57\meter,\, 3.32\degree$ & $\mathbf{0.631\meter},\, 3.906\degree$ \\
\noalign{\smallskip}\hline\noalign{\smallskip}
Average & $2.08\meter,\, 6.83\degree$ & $1.92\meter,\, 6.28\degree$ & $1.33\meter,\, 5.17\degree$ & $1.30\meter,\, 5.52\degree$  & $1.62\meter,\, 2.86\degree$ & $\mathbf{0.784\meter,\, 2.817\degree}$ \\
\noalign{\smallskip}\hline\noalign{\smallskip}
\end{tabular}
\vspace{-0.4cm}
\end{table*}

We compare the performance of our VLocNet architecture with current state-of-the-art deep learning-based localization methods namely PoseNet~\cite{kendall2015convolutional}, Bayesian PoseNet~\cite{kendall2015modelling}, LSTM-Pose~\cite{walch16spatialstms}, VidLoc~\cite{Clark2017VidLoc6V}, Hourglass-Pose~\cite{MelekhovYKR17}, BranchNet~\cite{wu2017}, PoseNet2~\cite{kendall2017geometric}, SVS-Pose~\cite{naseer17iros}, and NNnet~\cite{laskar2017camera}. We report the performance in terms of the median translation and orientation errors for each scene in the datasets. On the 7-Scenes dataset, we initialized the $\hat{s}_x$ and $\hat{s}_q$ for our loss function with values between ${-3}$ to $0$ and ${-4.8}$ to ${-3}$ respectively. \tabref{tab:7scenesCompGP} reports the comparative results on this dataset. Our VLocNet architecture consistently outperforms the state-of-the-art methods for all the scenes by $77.14\%$ in translation and $59.14\%$ in rotation. On the Cambridge Landmarks dataset, we report the results using $\hat{s}_x=-3$ and $\hat{s}_q=-6.5$ for all the scenes. Using our VLocNet architecture with the proposed Geometric Consistency Loss, we improve upon the current state-of-the-art results by $51.6\%$ in translation and $1.5\%$ in orientation. Note that we did not perform any hyperparameter optimization, we expect further improvements to the results presented here by tuning the parameters. The results demonstrate that our network substantially improves upon the state-of-the-art on both indoor as well as outdoor datasets.

In order to evaluate the performance of VLocNet on visual odometry estimation, we show quantitative comparison against three state-of-the-art CNN approaches, namely DeepVO~\cite{MohantyDeepVO}, cnnBspp~\cite{melekhov2017cnnBspp} and LBO~\cite{NicolaiRSSw2016}. \tabref{tab:7scenesCompVO} shows comprehensive results from this experiment on the 7-Scenes dataset. For each scene, we report the average translational and rotational error as a function of sequence length. As illustrated in \tabref{tab:7scenesCompVO}, our network achieves an improvement of $27.0\%$ in translation and $16.67\%$ in orientation outperforming the aforementioned approaches and thus reinforcing its suitability for visual odometry estimation.

\begin{table}
\footnotesize 
\centering
\caption{Comparison of 6DoF visual odometry on the 7-Scenes dataset.}
\label{tab:7scenesCompVO}
\begin{tabular}{p{1.2cm}p{1.2cm}p{1.3cm}p{1.45cm}|p{1.2cm}}
\hline\noalign{\smallskip}
Scene & LBO~\cite{NicolaiRSSw2016} & DeepVO \cite{MohantyDeepVO} & cnnBspp \cite{melekhov2017cnnBspp} & VLocNet (Ours) \\
\noalign{\smallskip}\hline\hline\noalign{\smallskip}
Chess & $1.69,\,1.13$ & $2.10,\,1.15$ & $1.38,\,1.12$ & $\mathbf{1.14,\,0.75}$ \\
Fire & $3.56,\,1.42$ & $5.08,\,1.56$ & $2.08,\,1.76$ & $\mathbf{1.81},\,1.92$ \\
Heads & $14.43,\,2.39$ & $13.91,\,2.44$ & $3.89,\,2.70$ & $\mathbf{1.82,\,2.28}$ \\
Office & $3.12,\,1.92$ & $4.49,\,1.74$ & $1.98,\,1.52$ & $\mathbf{1.71,\,1.09}$ \\
Pumpkin & $3.12,\,1.60$ & $3.91,\,1.61$ & $1.29,\,1.62$ & $\mathbf{1.26,\,1.11}$ \\
RedKitchen & $3.71,\,1.47$ & $3.98,\,1.50$ & $1.53,\,1.62$ & $\mathbf{1.46,\,1.28}$ \\
Stairs & $3.64,\,2.62$ & $5.99,\,1.66$ & $2.34,\,1.86$ & $\mathbf{1.28,\,1.17}$ \\
\noalign{\smallskip}\hline\noalign{\smallskip}
Average & $4.75,\,1.79$ & $5.64,\,1.67$ & $2.07,\,1.74$ & $\mathbf{1.51,\,1.45}$ \\
\noalign{\smallskip}\hline\noalign{\smallskip}
\multicolumn{5}{r}{\footnotesize{translation [$\%$], orientation [$\degm$]}}
\end{tabular}
\vspace{-0.5cm}
\end{table}

\subsection{Benchmarking}
\label{sec:benchmarking}

As mentioned in previous works~\cite{walch16spatialstms}, no deep learning-based localization method thus far has been able to match the performance of state-of-the-art local feature-based approaches. In order to gain insights on the performance of VLocNet, we present results on the 7-Scenes dataset, in comparison with Active Search (without prioritization)~\cite{sattler2017}, which is a state-of-the-art SIFT-based localization method. Moreover, as a proof of validation that our trained network is able to regress poses beyond those shown in the training images, we also compare with Nearest Neighbor localization~\cite{kendall2015convolutional}. \tabref{tab:7scenesBench} shows the comparative results of VLocNet against the aforementioned methods. 

Local feature-based approaches often fail to localize in textureless scenes due to the inadequate number of correspondences found. In \tabref{tab:7scenesBench}, we denote the number of images for which the localization fails in parenthesis and for a fair comparison we report the average accuracy of Nearest Neighbor and Active Search for only the images that the localization succeeded. The results show that VLocNet outperforms the Nearest Neighbor approach by 90.20\% in translation and 70.98\% in orientation, in addition to having no localization failures. This is by far the largest improvement achieved by any CNN-based approach. Moreover, VLocNet achieves state-of-the-art performance in comparison to Active Search on four out of the seven scenes, in addition to achieving overall lower average translation error. Thus far, it was believed that CNN-based methods could be used complementary to SIFT-based approaches as they performed better in challenging perceptual conditions but in other cases they were outperformed by SIFT-based methods. We believe that these results have demonstrated the contrary and have shown that CNN-based approaches are not only more robust but also have the potential to outperform local feature-based methods. 

\begin{table}
\footnotesize 
\centering
\caption{Benchmarking median errors on the 7-Scenes dataset.}
\label{tab:7scenesBench}
\begin{tabular}{p{1.3cm}p{1.8cm}p{1.95cm}|p{1.7cm}}
\hline\noalign{\smallskip}
Scene & Nearest Neighbor~\cite{kendall2015convolutional} & Active Search~\cite{sattler2017} & VLocNet (Ours) \\
\noalign{\smallskip}\hline\hline\noalign{\smallskip}
Chess & $0.41\meter,\, 11.2\degree (0)$ & $0.04\meter,\, 1.96\degree (0)$ & $\mathbf{0.036\meter,\, 1.707\degree}$ \\
Fire & $0.54\meter,\, 15.5\degree (1)$ & $0.03\meter,\, 1.53\degree (1)$ & $0.039\meter,\, 5.338\degree$ \\
Heads & $0.28\meter,\, 14.0\degree (1)$ & $0.02\meter,\, 1.45\degree (1)$ & $0.046\meter, 6.645\degree$ \\
Office & $0.49\meter,\, 12.0\degree (34)$ & $0.09\meter,\, 3.61\degree (34)$ & $\mathbf{0.039\meter,\, 1.953\degree}$ \\
Pumpkin & $0.58\meter,\, 12.1\degree (68)$ & $0.08\meter,\, 3.10\degree (68)$ & $\mathbf{0.037\meter,\, 2.280\degree}$ \\
RedKitchen & $0.58\meter,\, 11.3\degree (0)$ & $0.07\meter,\, 3.37\degree (0)$ & $\mathbf{0.039\meter,\, 2.205\degree}$ \\
Stairs & $0.56\meter,\, 15.4\degree (0)$ & $0.03\meter,\, 2.22\degree (0)$ & $0.097\meter,\, 6.476\degree$ \\
\noalign{\smallskip}\hline\noalign{\smallskip}
Average & $0.49\meter,\, 13.1\degree$ & $0.05\meter,\, 2.46\degree$ & $\mathbf{0.048\meter},\, 3.801\degree$ \\
\noalign{\smallskip}\hline\noalign{\smallskip}
\multicolumn{4}{r}{\footnotesize{(n) localization failures}}
\end{tabular}
\vspace{-0.5cm}
\end{table}

\subsection{Architectural Analysis}
\label{sec:vlocEvaluation}

In this section, we quantitatively analyze the effect of the various architectural decisions made while designing \mbox{VLocNet}. Specifically, we show the performance improvements for the following:
\begin{itemize}
\item VLocNet-M1: ResNet-50 base architecture with ReLUs, $L^2$ Euclidean loss for translation and rotation with $\beta=1$
\item VLocNet-M2: ResNet-50 base architecture with ELUs, $L^2$ Euclidean loss for translation and rotation with $\beta=1$
\item VLocNet-M3: ResNet-50 base architecture with ELUs and previous pose fusion using $\mathcal{L}_{Geo}$ loss with $\beta=1$
\item VLocNet-M4: ResNet-50 base architecture with ELUs and previous pose fusion using $\mathcal{L}_{Geo}$ loss with $\hat{s}_x$, $\hat{s}_q$
\end{itemize}

\begin{figure}
\centering
\includegraphics[width=0.485\linewidth]{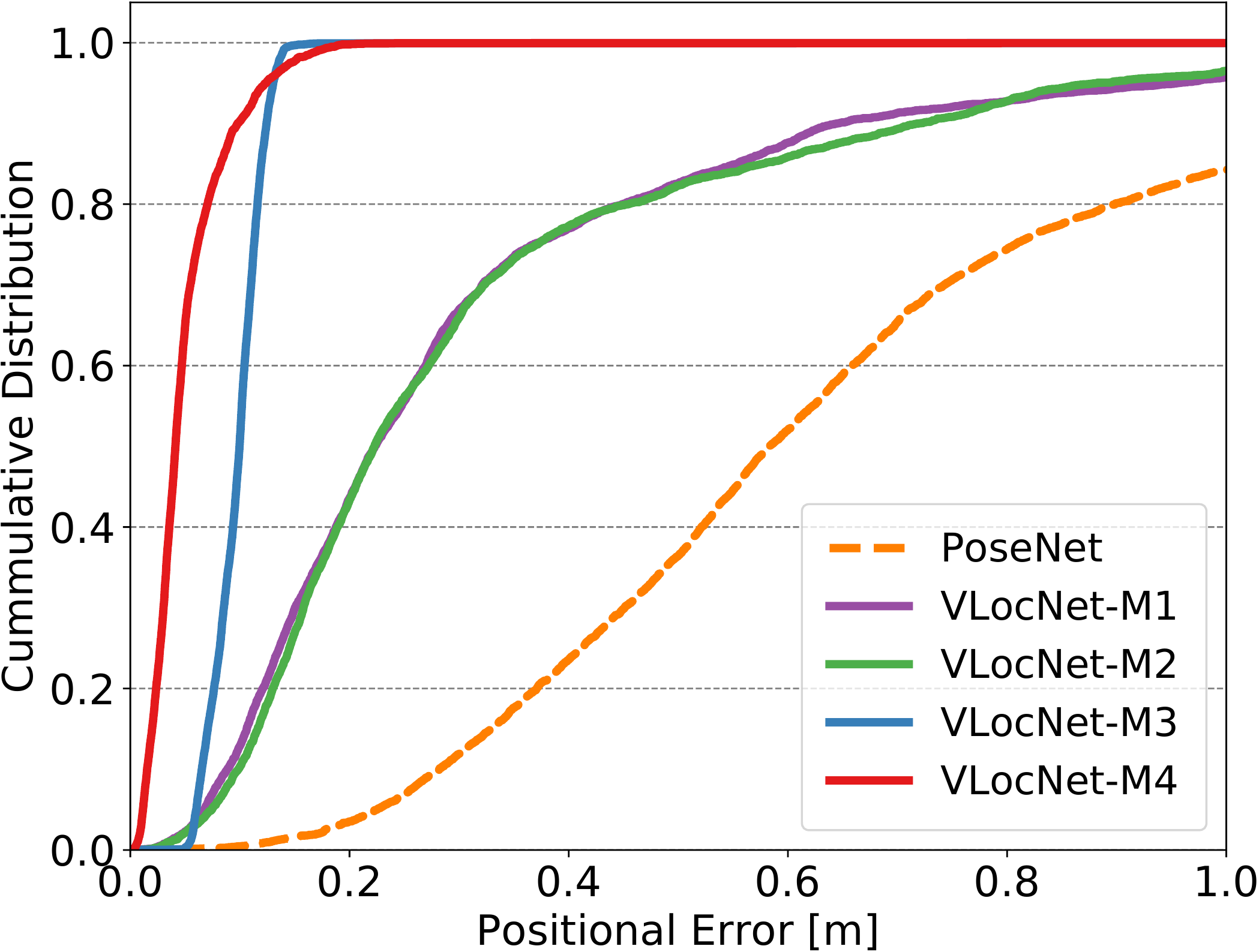}
\includegraphics[width=0.485\linewidth]{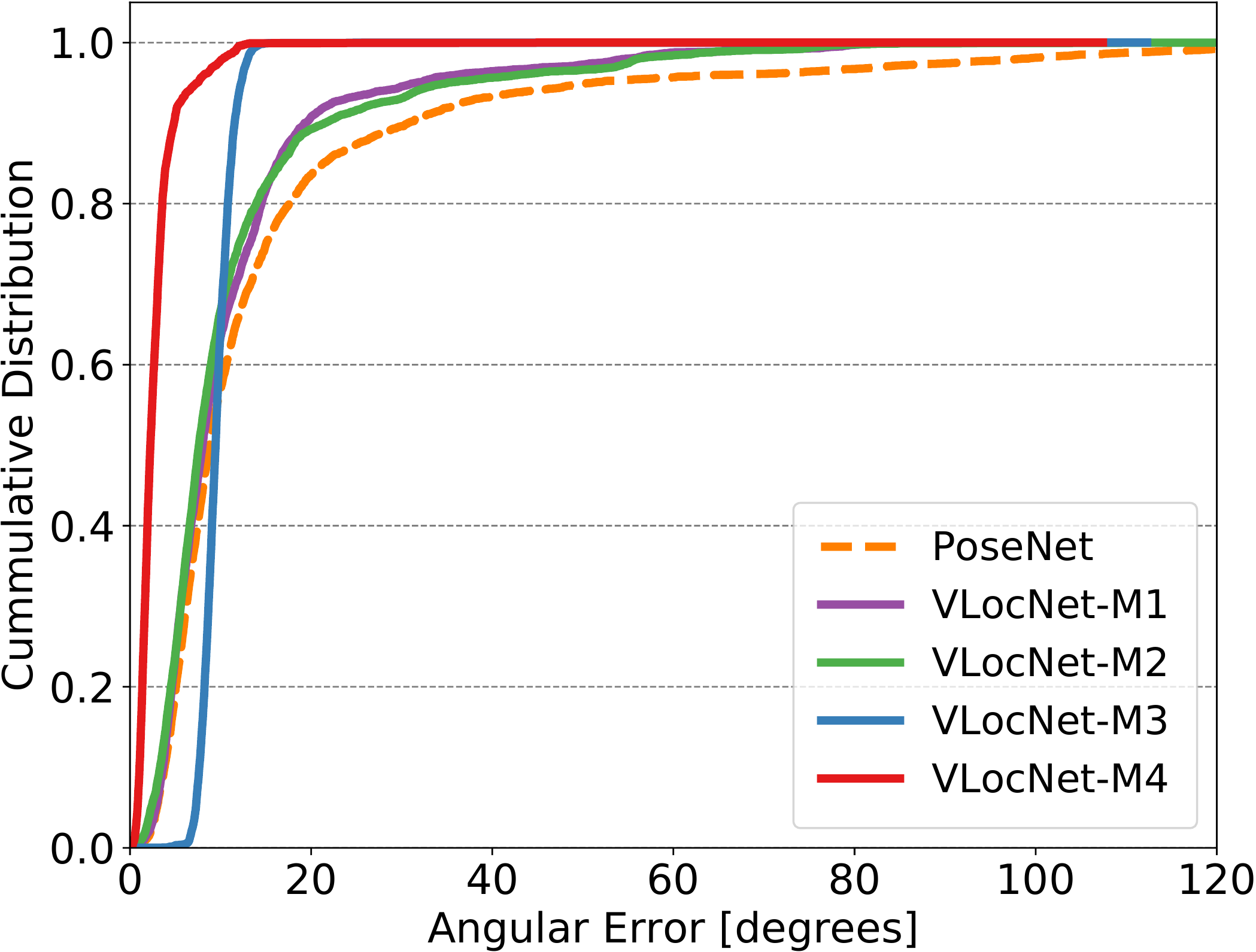}
\caption{Qualitative analysis of the localization performance for our proposed VLocNet architecture against PoseNet presented as a cumulative histogram of normalized errors for the Red Kitchen scene.}
\label{fig:cummlRedKitchen}
\vspace{-0.2cm}
\end{figure}

\tabref{tab:VLocNetAnalysis} shows the median error in pose estimation as an average of all the scenes in the 7-Scenes dataset. We observe that incorporating residual units in our architecture yields an improvement of 54.09\% and 14.68\% for the translation and orientation components respectively in comparison to PoseNet. However, the most notable improvement is achieved by fusing the previous pose information using our Geometric Consistency Loss, which can be seen in the improvement of the translational error between VLocNet-M2 and VLocNet-M3. This clearly shows that constricting the search space with the relative pose information while training substantially increases the performance. Furthermore, by utilizing learnable parameters for weighting the translational and rotational loss terms, our network yields a further improvement in performance compared to manually tuning the weighting. In~\figref{fig:cummlRedKitchen} we show the cumulative histogram error of the aforementioned models trained on the RedKitchen scene. It can be seen that even our base VLocNet model (VLocNet-M1) shows a significant improvement over the baseline method for the translational error. Moreover our final architecture (VLocNet-M4) achieves a rotational error below $10\degree$ for $100\%$ of the poses. 

\begin{figure*}
\centering
\includegraphics[width=0.4\linewidth]{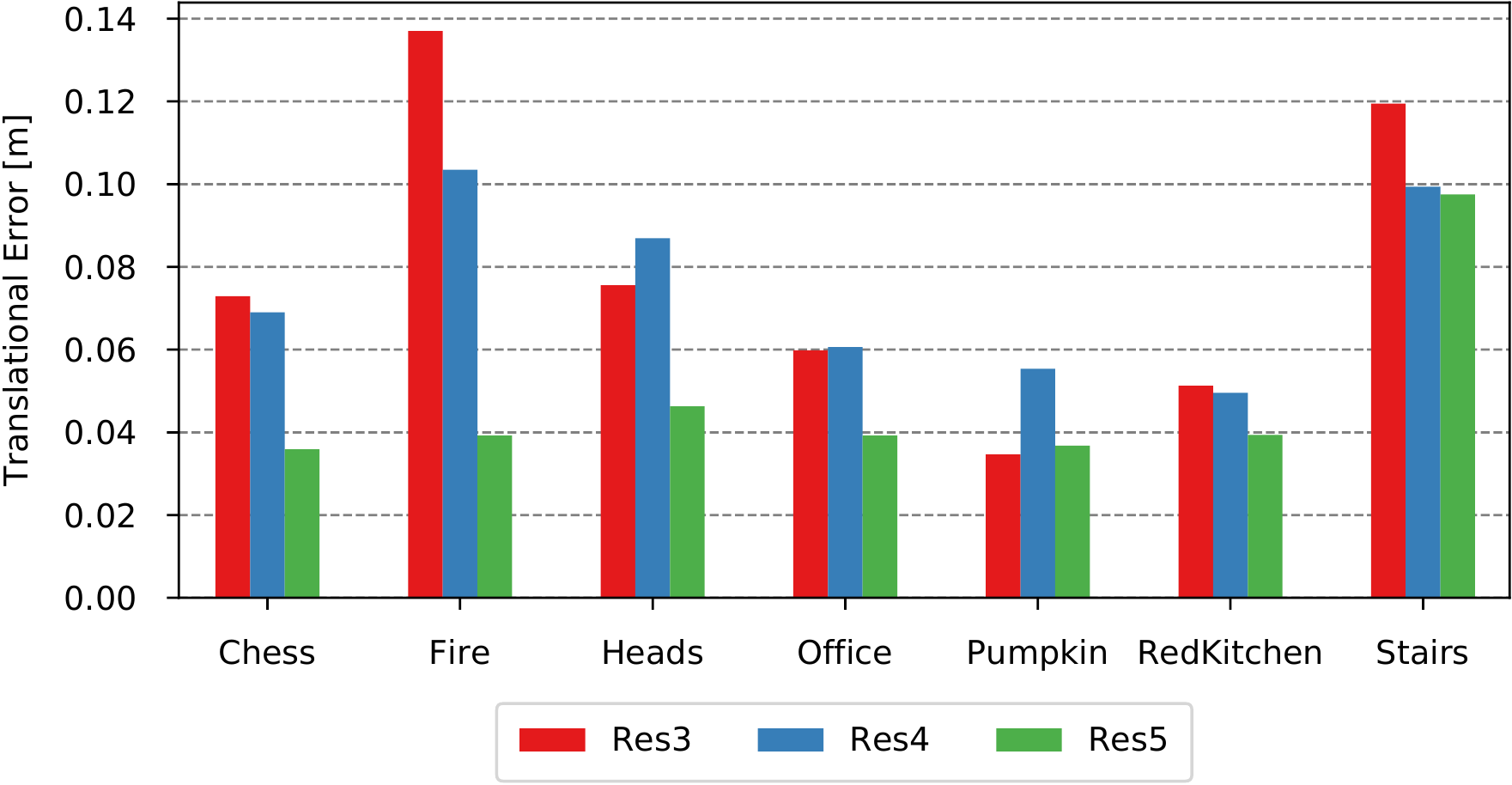} \qquad\qquad
\includegraphics[width=0.4\linewidth]{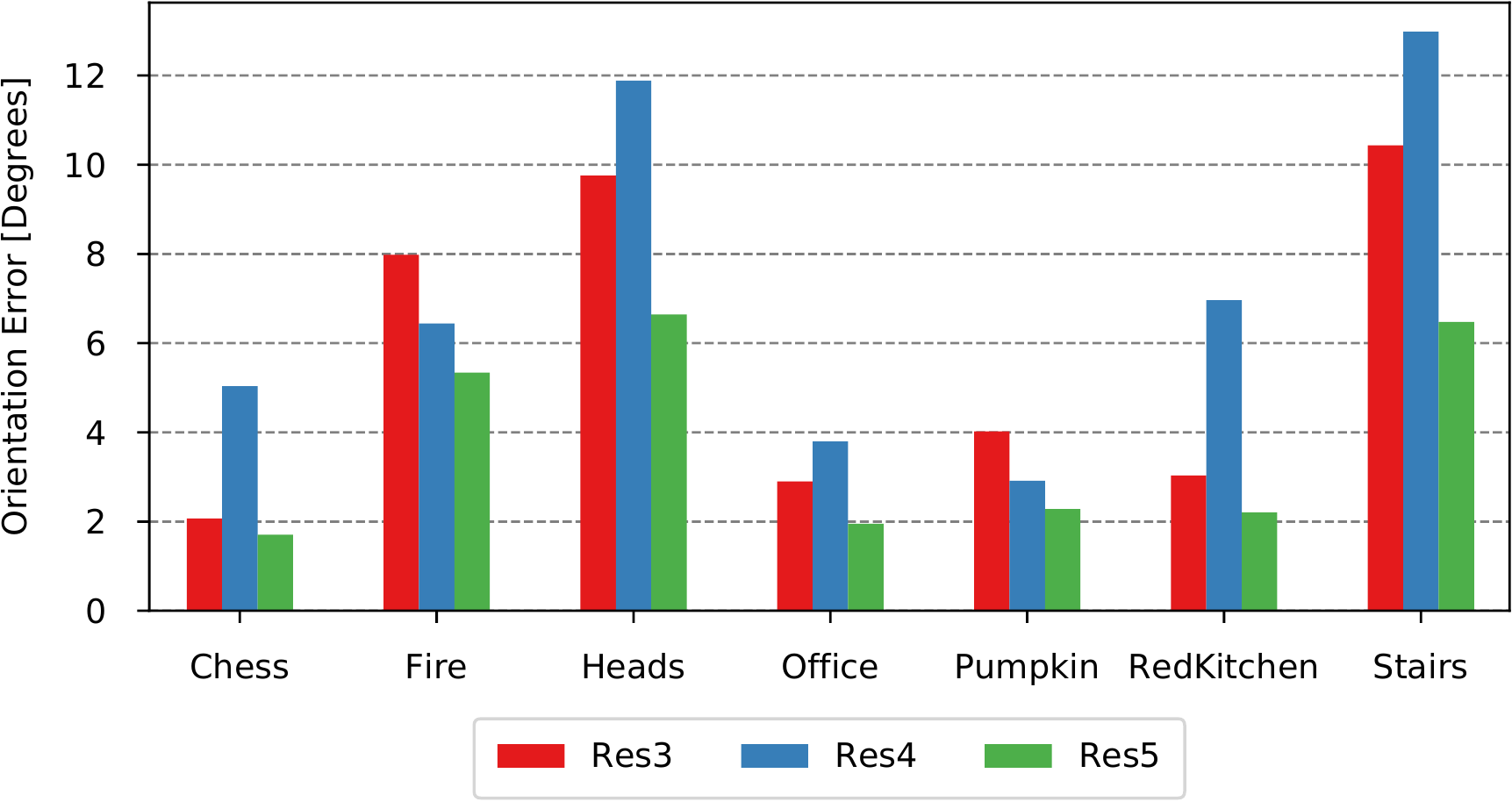}
\vspace{-0.2cm}
\caption{Comparison of median localization error from fusing previous pose information at various stages in our VLocNet architecture with our proposed Geometric Consistency Loss. The results consistently show that the highest localization accuracy is achieved by fusing the previous predicted pose at \textit{Res5}.}
\label{fig:prevPoseComp}
\vspace{-0.4cm}
\end{figure*}

We additionally performed experiments to determine the downsampling stage to fuse the previous pose while using our Geometric Consistency Loss function. \figref{fig:prevPoseComp} shows the median error while fusing the pose at \textit{Res3}, \textit{Res4} and \textit{Res5} in our architecture. It can be seen that fusing at \textit{Res5} where the feature maps are of size $7\times7$, yields the lowest localization error, while fusing at earlier stages produces varying results for different scenes; either lower translational error at the cost of the orientation or vice versa.

\begin{table}
\footnotesize 
\centering
\caption{Comparative analysis of VLocNet on the 7-Scenes dataset.}
\label{tab:VLocNetAnalysis}
\begin{tabular}{p{2cm}p{1.3cm}p{1.3cm}}
\hline\noalign{\smallskip}
Model & Position & Orientation \\
\noalign{\smallskip}\hline\hline\noalign{\smallskip}
PoseNet~\cite{kendall2015convolutional} & $0.44\meter$ & $10.4\degree$ \\
VLocNet-M1 & $0.202\meter$ & $8.873\degree$ \\
VLocNet-M2 & $0.197\meter$ & $8.209\degree$ \\
VLocNet-M3 & $0.081\meter$ & $7.860\degree$ \\
VLocNet-M4 & $\mathbf{0.048\meter}$ & $\mathbf{3.801\degree}$ \\
\noalign{\smallskip}\hline\noalign{\smallskip}
\end{tabular}
\end{table}

\subsection{Evaluation of Deep Auxiliary Learning}
\label{sec:auxEvaluation}

\begin{figure}
\centering
\includegraphics[width=0.8\linewidth]{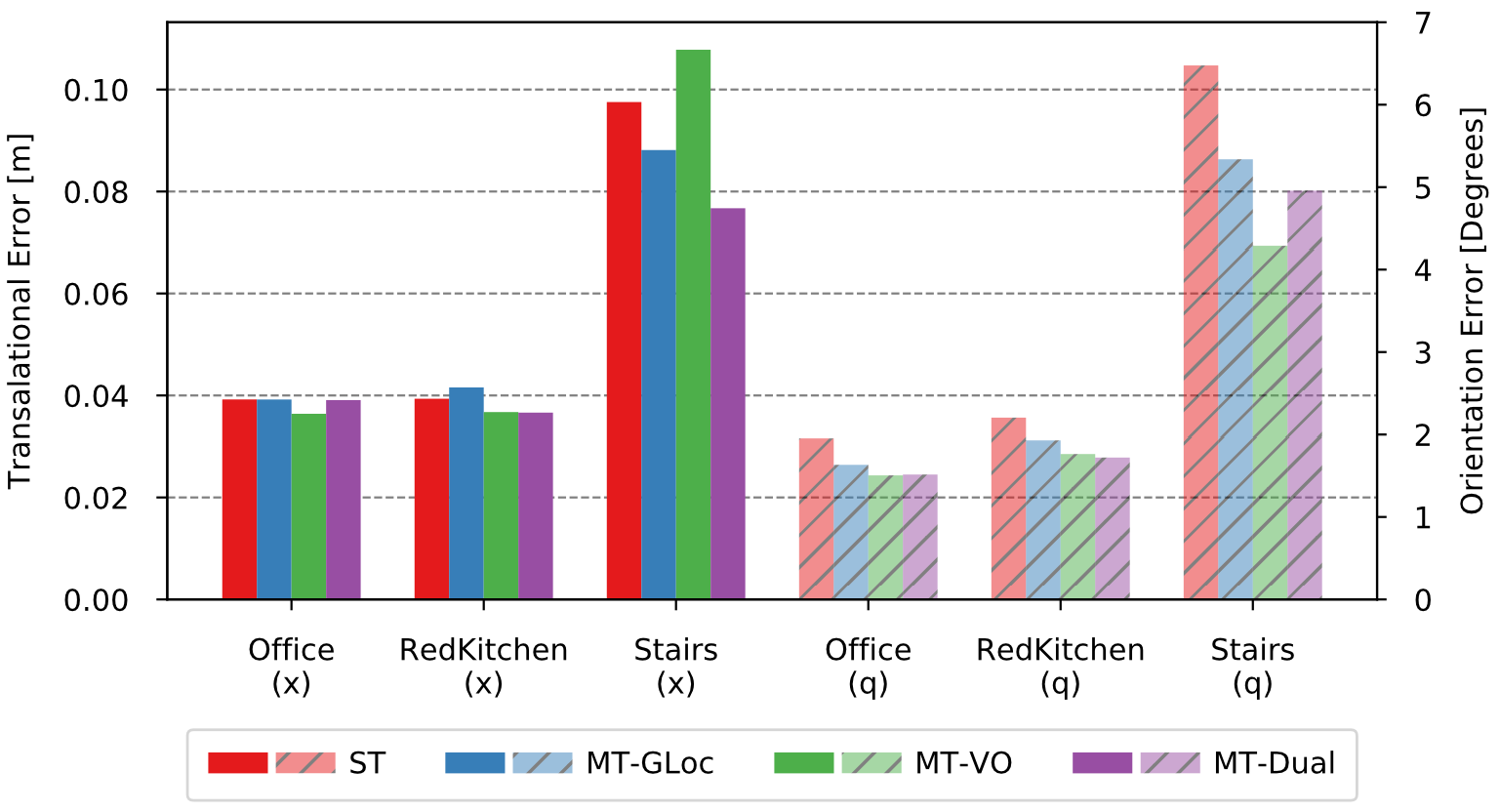}
\vspace{-0.2cm}
\caption{Performance of our single-task model in comparison to the multitask VLocNet with different weight initializations, on the 7-Scenes dataset. (x) and (q) denote the translation and orientation components.}
\label{fig:auxComp}
\end{figure}

In this section, we evaluate the performance of our jointly trained model using auxiliary learning, along with different optimization strategies that we employed. We explored using both joint and alternating optimization to minimize the loss. We found that the average localization error using an alternate optimization strategy was $28.99\%$ and $18.47\%$ lower in translation and rotation respectively, when compared to a joint optimization. This can be attributed to the difference in scales of the loss values for each task, resulting in the optimizer becoming more biased towards minimizing the global pose regression error at the cost of having suboptimal relative pose estimates. This inadvertently results in worse accuracy for both tasks.

When both global pose regression and visual odometry networks are trained independently, each of them alter the weights of their convolution layers in different ways. Therefore, we evaluated strategies that enable efficient sharing of features between both networks to facilitate the learning of inter-task correlations. Using the model trained on our single-task global pose sub-network (ST) as a baseline, we evaluate the effect of different initializations of the joint model on the localization accuracy. More precisely, we compare the effect of initializing our VLocNet model using weights from: the pretrained task-specific global pose network \mbox{(MT-GLoc)}, the pretrained task-specific visual odometry network \mbox{(MT-VO)}, and the combined weights from both networks \mbox{(MT-Dual)}. \figref{fig:auxComp} shows the results from this experiment. It can be seen that our joint models that regress relative poses as an auxiliary task, outperform each of the task-specific models, demonstrating the efficacy of our approach. The improvement is most apparent in the Stairs scene which is the most challenging scene in the 7-Scenes dataset as it contains repetitive structures and textureless surfaces. Furthermore, on closer examination, we find that dual initialization of both sub-networks with weights from their task-specific models achieves the best performance, contrary to initializing only one of the sub-networks and learning the other from scratch. Another interesting observation worth noting is that initializing only the global localization stream in the joint network with pretrained weights yields the lowest improvement in pose accuracy compared to the single-task model. This is to be expected as the visual odometry stream does not provide reasonable estimates when the training begins, therefore the localization stream cannot benefit from the motion specific features from the odometry stream.

We summarize the localization performance achieved by our multitask VLocNet incorporating the Geometric Consistency Loss while simultaneously training the auxiliary odometry network in \tabref{tab:shareComp}, where we vary the number of shared layers between the global localization and the visual odometry streams. The table shows the median pose error as an average over all the 7-scenes. We experimented with maintaining a shared stream up to the end of \textit{Res2}, end of \textit{Res3} and end of \textit{Res4} in our architecture. The results indicate that the lowest average error is achieved by sharing the streams up to \textit{Res3}, which shows that features learned after \textit{Res3} are highly task-specific and features learned before \textit{Res2} are too generic. In comparison to the single-task \mbox{VLocNet} model, the multitask variant achieves an improvement of $12.5\%$ in translational and $18.49\%$ in rotational components of the pose. We believe that these results demonstrate the utility of learning joint multitask models for visual localization and odometry. A live online demo can be viewed at \protect\url{http://deeploc.cs.uni-freiburg.de/}.

\begin{table}
\footnotesize 
\centering
\caption{Summary of the localization performance achieved by VLocNet with varying amounts of sharing.}
\label{tab:shareComp}
\begin{tabular}{p{1.7cm}p{1.6cm}p{1.7cm}p{1.7cm}}
\hline\noalign{\smallskip}
& Res2 & Res3 & Res4 \\
\noalign{\smallskip}\hline\hline\noalign{\smallskip}
7-Scenes Avg. & $0.055\meter,\, \mathbf{2.989\degree}$ & $\mathbf{0.042\meter},\, 3.098\degree$ & $0.053\meter,\, 3.174\degree$ \\
\noalign{\smallskip}\hline\noalign{\smallskip}
\end{tabular}
\vspace{-0.4cm}
\end{table}

\section{Conclusion}
\label{sec:conclusion}

In this paper, we proposed a novel end-to-end trainable multitask DCNN architecture for 6-DoF visual localization and odometry estimation from subsequent monocular images. We present a framework for learning inter-task correlations in our network using an efficient sharing scheme and a joint optimization strategy. We show that our jointly trained localization model outperforms task-specific networks, demonstrating the efficacy of learning visual odometry as an auxiliary task. Furthermore, we introduced the Geometric Consistency Loss function for regressing 6-DoF poses consistent with the true motion model.

Using extensive evaluations on standard indoor and outdoor benchmark datasets, we show that both our single-task and multitask models achieve state-of-the-art performance compared to existing CNN-based approaches, which accounts for an improvement of $80\%$ and $66.69\%$ in translation and rotation respectively. More importantly, our approach is the first to close the performance gap between local feature-based and CNN-based localization methods, even outperforming them in some cases. Overall, our findings are an encouraging sign that utilizing multitask DCNNs for localization and odometry is a promising research direction. As future work, we plan to investigate joint training with more auxiliary tasks such as semantic segmentation and image similarity learning that can further improve the performance.

%% Use plainnat to work nicely with natbib. 
\bibliographystyle{IEEEtrans}
\footnotesize
%\balance
\bibliography{references}


%%%%%%%%%%%%%%%%%%%%%%%%%%%%%%%%%%%%%%%%%%%%%%%%%%%%%%%%%%%%%%%%%%%%%%%%%%%%%%%%

%%%%%%%%%%%%%%%%%%%%%%%%%%%%%%%%%%%%%%%%%%%%%%%%%%%%%%%%%%%%%%%%%%%%%%%%%%%%%%%%

%%%%%%%%%%%%%%%%%%%%%%%%%%%%%%%%%%%%%%%%%%%%%%%%%%%%%%%%%%%%%%%%%%%%%%%%%%%%%%%%

%%%%%%%%%%%%%%%%%%%%%%%%%%%%%%%%%%%%%%%%%%%%%%%%%%%%%%%%%%%%%%%%%%%%%%%%%%%%%%%%

\end{document}